\documentclass{article}
\usepackage[utf8]{inputenc}
\usepackage[T1]{fontenc}
\usepackage[english]{babel}
\usepackage{amsmath}
\usepackage{amssymb,amsfonts,textcomp}
\usepackage{array}
\usepackage{supertabular}
\usepackage{hhline}
\usepackage[pdftex]{graphicx}
\makeatletter
\newcommand\arraybslash{\let\\\@arraycr}
\makeatother
\begin{document}
\title{On the Convergence of SGD Training of Neural Networks}
\author{Thomas M. Breuel \\ Google, Inc. \\ {\tt tmb@google.com}}
\date{}
\maketitle

\begin{abstract}
Neural networks are usually trained by some form of stochastic gradient
descent (SGD)). A number of strategies are in common use intended to
improve SGD optimization, such as learning rate schedules, momentum, and
batching. These are motivated by ideas about the occurrence of local
minima at different scales, valleys, and other phenomena in the
objective function. Empirical results presented here suggest that these
phenomena are not significant factors in SGD optimization of MLP-related
objective functions, and that the behavior of stochastic gradient
descent in these problems is better described as the simultaneous
convergence at different rates of many, largely non-interacting
subproblems.
\end{abstract}

\section{Introduction}

An one hidden layer multi-layer perceptron (MLP) is a nonlinear
discriminant function

$$ y = f_\theta(x) = \sigma(b+B\cdot\sigma(a+A\cdot x)) $$

Here, the parameters  $\theta$ are
two vectors  $a,b$  and two
matrices  $A,B$ , and 
$\sigma(...)$  is the element-wise
sigmoid function of a vector. 

These parameters are usually optimized by
stochastic gradient descent, minimizing some objective function
involving  $y$  and the
ground truth classification information. Let us call the objective
function  $C(\theta)$ .
Much effort has been invested in coming up with efficient ways of
performing this optimization or stochastic gradient 
descent\cite{sgd1,sgd2,sgd3,sgd4,so1}.

Generally,  $f_\theta$ is thought to
be a function that has many local minima.
Indeed, local minima are frequently observed
in training small neural networks on problems like XOR,
parity, and spiral problems.
Much effort in the literature has been expended on
ways to train despite the existence of such local
minima\cite{min1,min2,min3}.

Furthermore, any single local
minimum gives rise to many more local minima because of symmetries of 
$f_\theta$ in its parameter vector
(e.g., corresponding permutations of the rows of 
$A$  and the columns of 
$B$  yield the same
classifier). Commonly,  $C(\theta)$
is also thought to have a hierarchy of minima at different scales.
These assumptions have yielded a number of common strategies for neural
network training. For example, momentum is intended to smooth out local
minima at small scales, and learning rate schedules are supposed to
cope with a hierarchy of minima at different scales, in a way similar
to simulated annealing.

Although these assumptions about the objective function are commonly
made, they have received little empirical testing. That is in part
because observing the convergence of MLPs in weight space is difficult
because many different weight settings give equivalent outputs. In
addition, it is difficult to explore the very high dimensional weight
space of MLPs.

Here, we examine the question of how stochastic gradient descent
optimization proceeds for MLPs with two simple approaches:

\begin{itemize}
\item We trace  $C(\theta)$ over
time as optimization proceeds for multiple different starting points.
\item Instead of comparing the 
$\theta$ parameters, which are
beset by complicated symmetries, we use a collection of test set
samples  $\{\xi_1, ... \xi_N\}$  and compute
the corresponding vector 
$\tau(\theta) = (f_\theta(\xi_1),...,f_\theta(\xi_N))$. For two weight
vectors  $\theta$  and $\theta'$  to be similar under
symmetries, it is necessary (but not sufficient) for 
$\tau(\theta)\approx\tau(\theta')$. 
\end{itemize}

If the standard view of stochastic gradient descent (SGD) optimization
of neural networks is correct, i.e., if there are numerous local minima,
then the following predictions would hold true:

\begin{itemize}
\item For a given learning rate, SGD optimization would continue until
the learning rate is too large and the search simply moves around in a
random walk within the basin surrounding the local minimum.
\item For a given learning rate, the parameter space 
of $\theta$ should be partitioned
into regions, each of which corresponds to the basin surrounding a
local minimum at the scale corresponding to the learning rate.
\item We should be able to find multiple different starting points in
parameter space that converge to the same basin surrounding a local
minimum at a particular scale.
\end{itemize}

Note that for these predictions, the necessary-but-not-sufficient aspect
of  $\tau$-equivalence of
parameters is not a problem. That is, the fact that 
$\tau$-equivalence may
count some parameter vectors as equivalent that are, in fact, not
equivalent does not invalidate any of these predictions; they hold true
under both  $\tau$-equivalence of parameters and true equivalence.

In fact, there are two different notions of parameter equivalence we use
in the experiments below. The first is 
$\tau$-equivalence as
defined above, namely a vector of discriminant function values
predicted by the neural network on the test set. The second, let’s call
it $\kappa$-equivalence
replaces the discriminant functions with the class predictions for each
test set sample. Under that weaker notion of equivalence, two sets of
parameters are equivalent if they make the same classifications on the
test set.

All the experiments reported below were carried out using the Torch
machine learning library on the deskewed MNIST dataset using a one
hidden layer MLP with 100 hidden units. The first 1000 samples of the
official test set were used as the test set for predicting the 
$\tau$  and 
$\kappa$ vectors. All
dimensionality reductions use the multidimensional scaling
implementation from scikits-learn.

\section{Large Scale Convergence}

\begin{figure}[thp]
 \includegraphics[width=\textwidth]{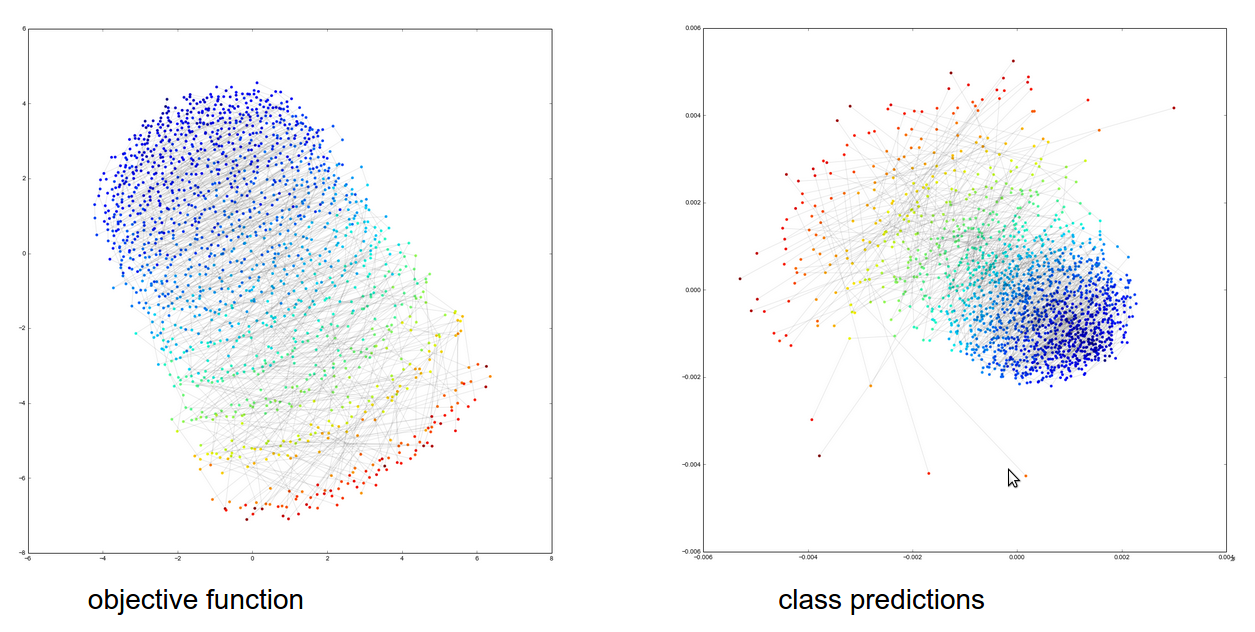} 
\caption{\label{initial}
$\tau$ and $\kappa$ predictions for
100 networks over time using multidimensional scaling. Training epochs
are indicated by color (from red to blue).
}
\end{figure}

The initial experiment consisted of training 100 networks over 20
epochs. The results are shown in Figure~\ref{initial}. We can see that both
discriminant functions and classification vectors over the test set
start off in one region of space and then generally converge towards
each other. The fact that the initial 
$\tau$ vectors appear to be
much more diverse than the initial 
$\kappa$ vectors is a result of
the nonlinearity implied by the argmax function that converts the
discriminant function into classifications. That is, during the first
epoch, the predictions $f_\theta(x)$
are all close to 0.5, but the use of the argmax function converts these
very similar discriminant function values into wildly different
classification results.

The natural next step is to visualize the space of paths after the
effects of the initial training have worn off. This is shown in Figure
2. At first glance, this figure appears to show a large number of local
minima, with each network stuck inside its own local minimum. However,
in fact, what this image shows is a top down projection of a number of
SGD paths that are all converging towards the center.

\begin{figure}[thp]
\includegraphics[width=\textwidth]{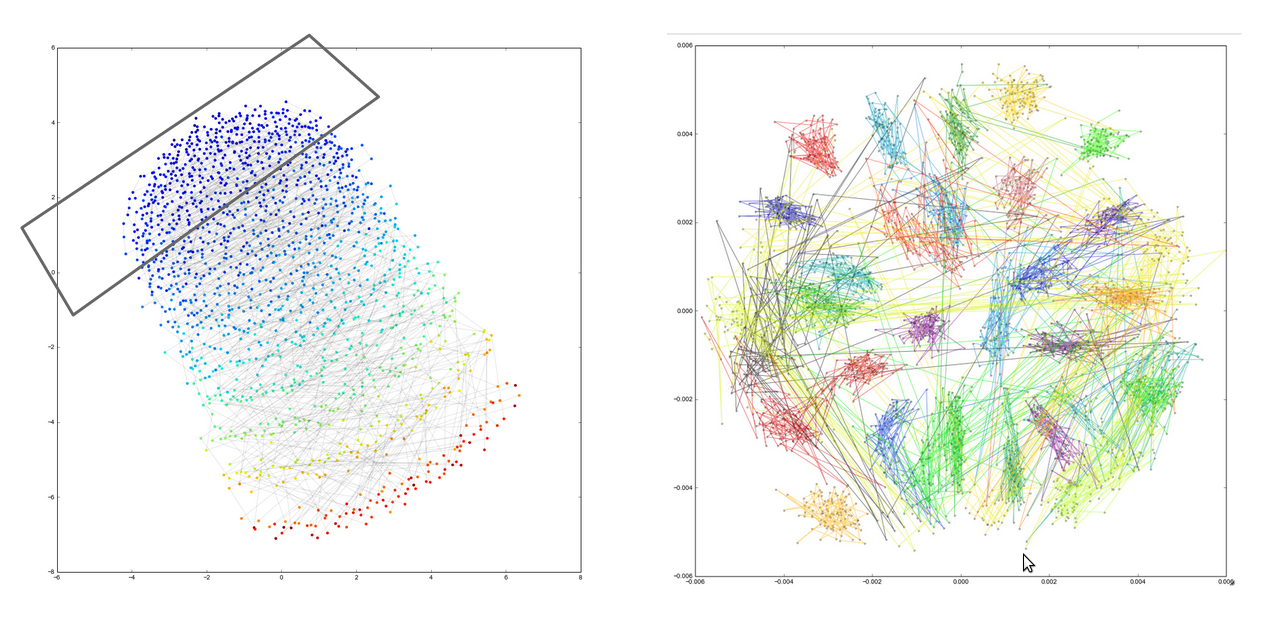}
\caption{\label{late}
Paths in parameter space after initial training. At first
glance, this appears to show multiple searches in local minima, but
actually, each of the apparent local minima represents a path that
slowly converges towards the center of the figure.
}
\end{figure}

We can see this much more easily if we follow a smaller number of SGD
paths, as shown in Figure 3. Now it is obvious that the paths are not,
in fact, stuck in local minima, but that they are still all making
progress towards a common optimum. To help with understanding the right
panel of Figure 2, the equivalent perspective is shown on the right in
Figure 3.

\begin{figure}[thp]
 \includegraphics[width=\textwidth]{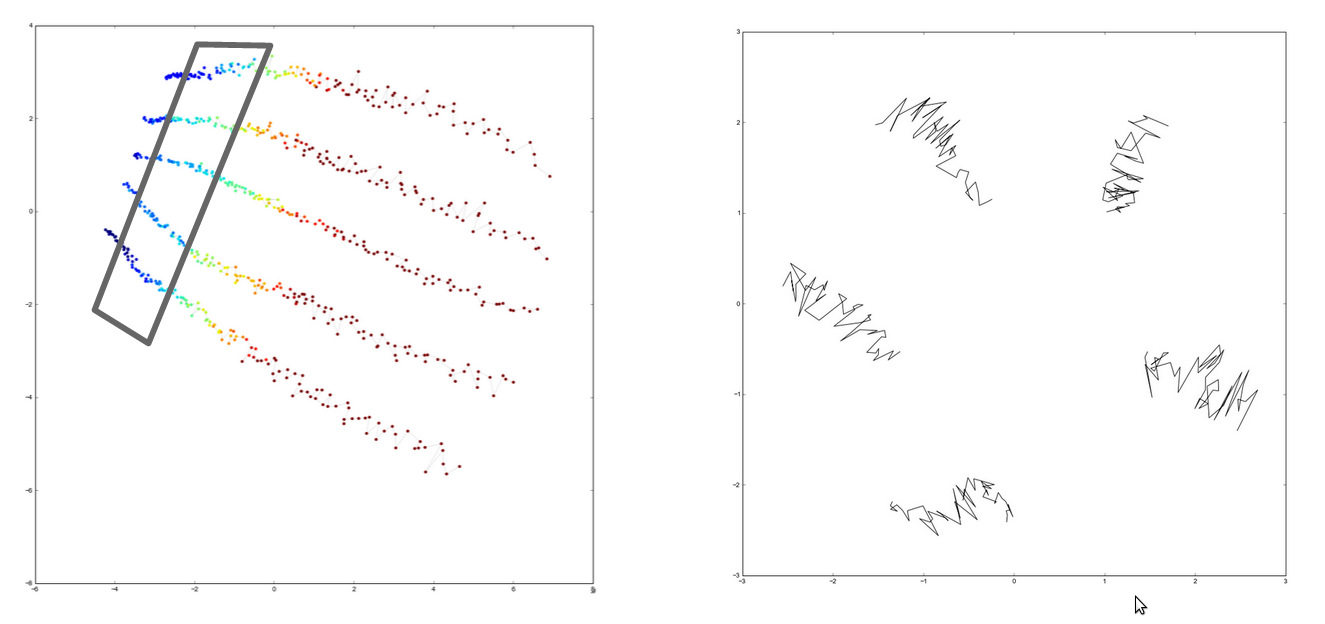} 
\caption{\label{5paths}
A small number of paths analyzed in the same way as Figure 2.
Now it is clear that none of the paths are stuck in local minima, and
instead keep making progress towards a common minimum near the top left
of the plot. The right hand side shows a slice through these paths,
which are now converting roughly towards the center.
}
\end{figure}

We find the same result if we look at the $\kappa$ function instead of the $\tau$ function.

\begin{figure}[thp]
\includegraphics[width=\textwidth]{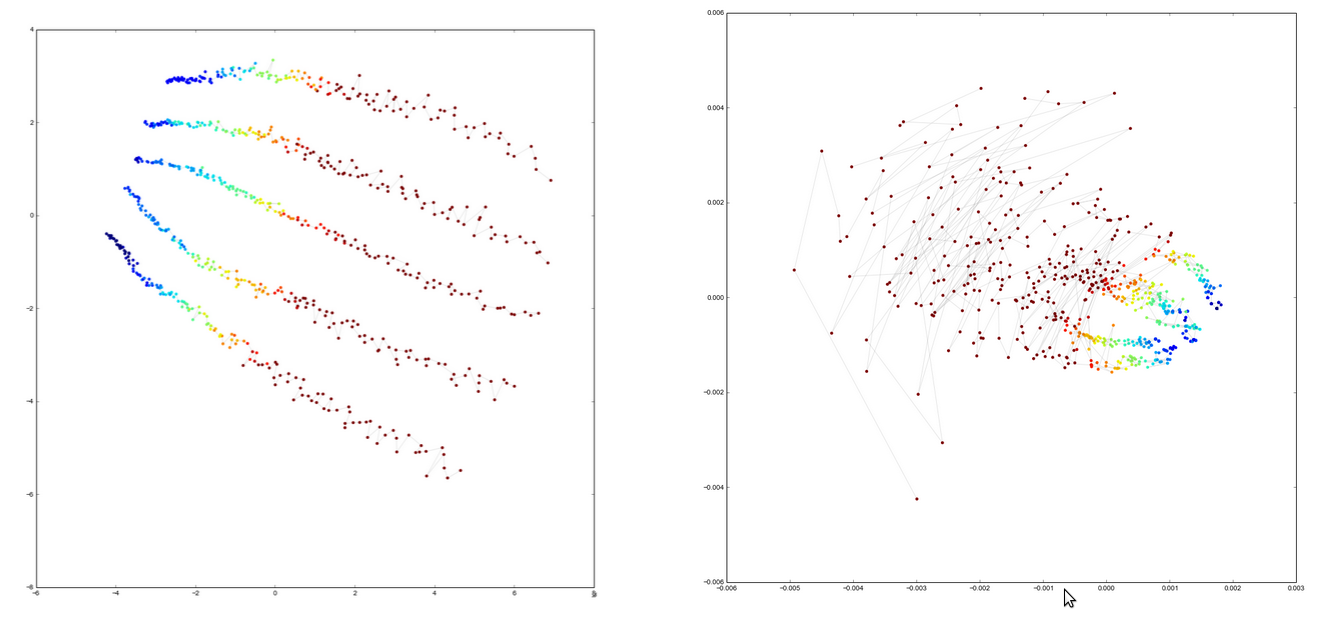}
\caption{\label{5paths-k}
The same analysis as in Figure 3, but using class labels (
$\tau$-equivalence) instead
of discriminant function values.
}
\end{figure}

The above experiments do not suggest the existence of local minima. In
fact, SGD training proceeds towards a common final minimum smoothly and
steadily along separate paths; each path “remembers” the initial
conditions, the random set of training weights that it was originally
trained with. There is no sign of local minima at all. This is quite
amazing if we recall that the classification error across these figures
goes from around 90\% to below 1\%.

To push these experiments to their extremes, Figure 6 shows the 
$\tau$-functions of the
parameters for epochs 50-500 for two networks, using a learning rate of
5 (a very high learning rate). Again, we see that each network steadily
learns, but makes slower and slower progress as time goes on. There is
no evidence of getting stuck in local minima, no evidence of minima at
different scales (which would show up as “bouncing around” in weight
space). Furthermore, the separation of the two networks in parameter
space is large compared to the size of the updates. Even over 500
epochs of training and at a final error rate of around 0.15\%, both
networks retain a memory of the initial weight choices.

\begin{figure}[thp]
 \includegraphics[width=\textwidth]{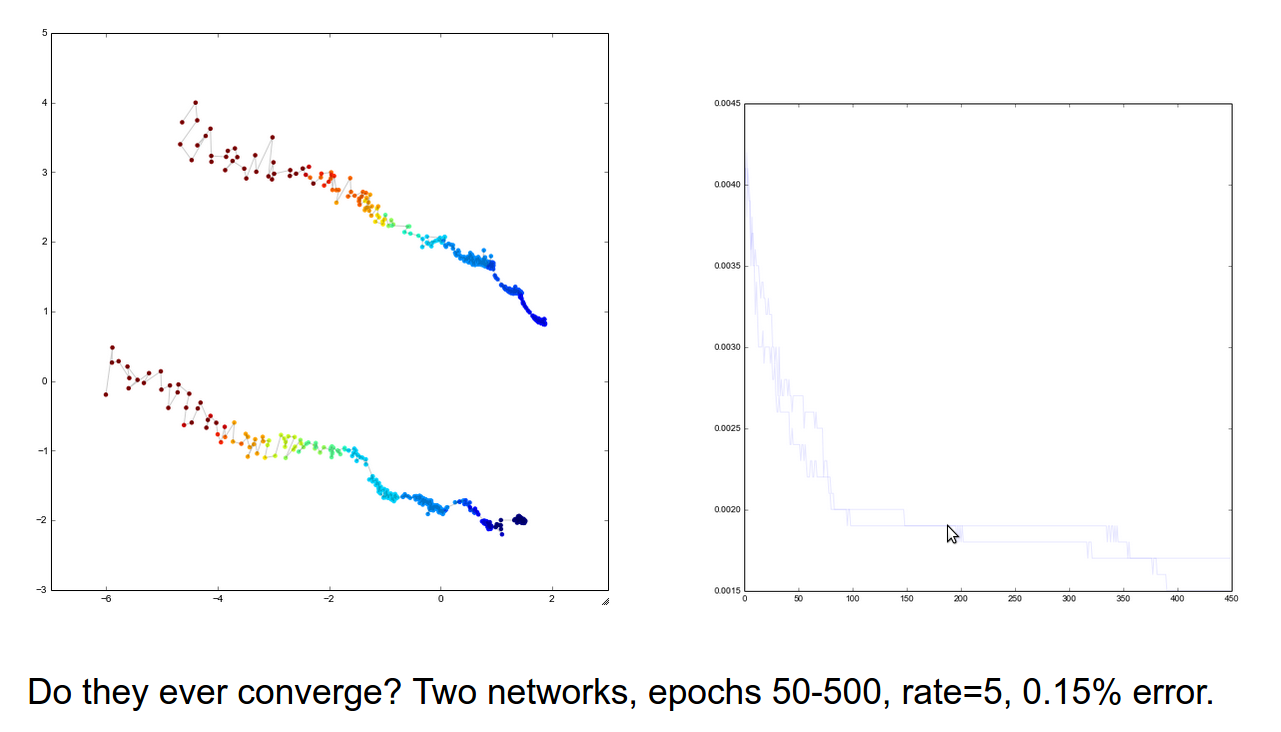}
\caption{\label{longtime}
Pushing two networks out over very long periods (epochs
50-500, learning rate = 5, final training set error is 0.15\%). The
networks retain a memory of their initial conditions over 500 epochs
and never reach a local minimum, common or otherwise, despite having
very similar classification error rates.
}
\end{figure}

One of the mechanisms proposed for improving convergence of stochastic
gradient descent is smoothing via momentum of batching. The effect of
this can be seen in Figure 6. We can see that batched updates do indeed
lead to a much smoother optimization path initially, but in the long
run, both paths smooth out and converge gradually. There is no evidence
that batching or averaging avoids “bouncing around in local minima” or
helps the MLP converge faster or better.

\begin{figure}[thp]
 \includegraphics[width=\textwidth]{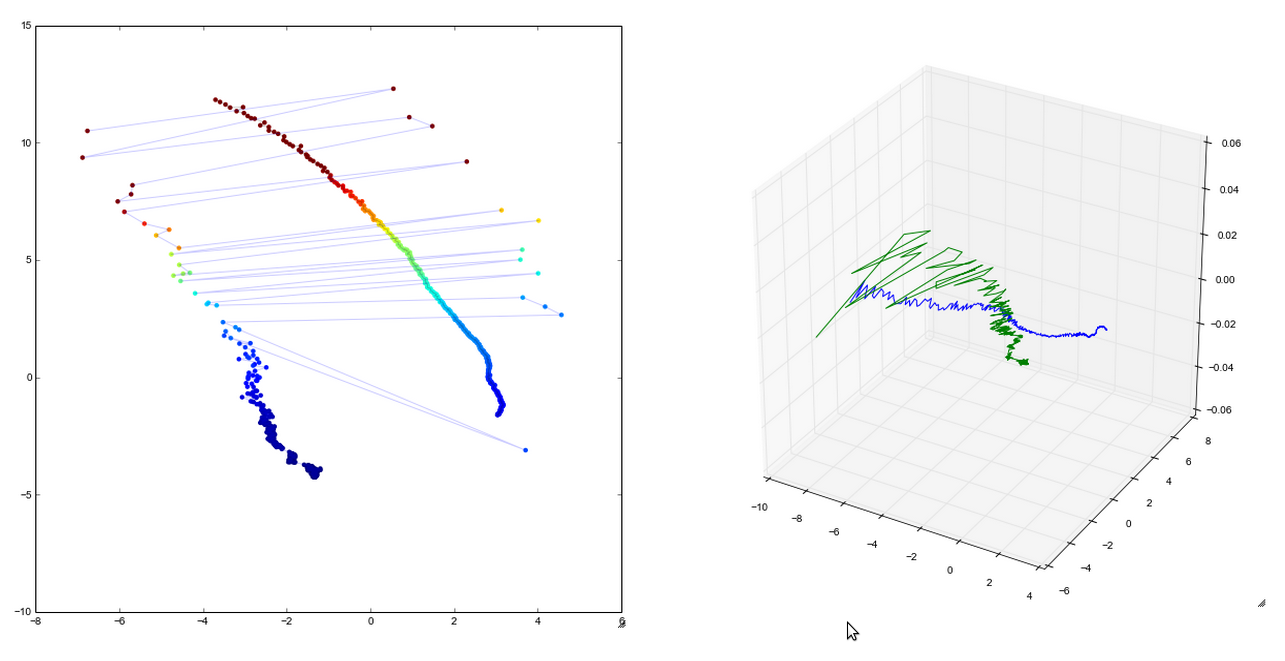} 
\caption{\label{batching}
Convergence of paths with and without batching. The smoother
path on the left and the blue path on the right use a batch size of
100, while the rougher path represents single sample update SGD
training. Although batching (or equivalently momentum) has the
predicted effect of smoothing out the training path, there is no
evidence that it improves the optimization process overall and training
progress appears to be similar for both cases.
}
\end{figure}

\section{A Stochastic Model}

The behavior of the SGD over time is quite odd, since the model retains
a memory of the initial conditions over very long time periods and
several orders of magnitude of the objective function. How is that
possible?

Consider a very simple SGD model, where we are simply training a vectors
$\theta$ to converge to 0. That
is, the objective function we are minimizing is 
$||\theta||^2$.
The SGD steps consist
of picking one of the dimensions of 
$\theta$ at random and
multiplying it with  $1-\epsilon$ 
for some small value of 
$\epsilon$. If we pick several
starting vectors for $\theta$ 
and “optimize” them with this procedure, we obtain the paths as shown
in Figure 7. This is the kind of behavior we might expect for SGD
optimization of MLPs: the memory of the initial conditions is quickly
erased and the paths convert to a common minimum (the global minimum in
this case).

\begin{figure}[thp]
 \includegraphics[width=\textwidth]{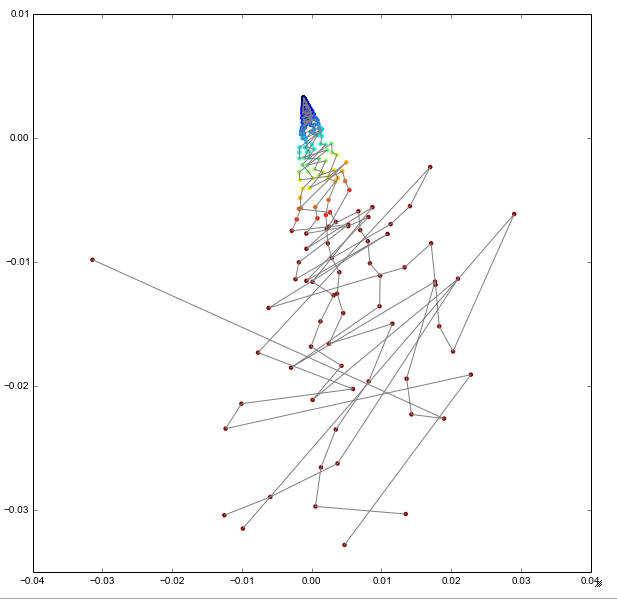}
\caption{\label{simple-sgd}
Simulated stochastic gradient descent of 5 random initial
1000-dimensional weight vector towards zero, after multidimensional
scaling. At each time step, a dimension is chosen uniformly at random
and multiplied by 0.9
}
\end{figure}

Figure 8 shows a slight variation of this, using an optimization
procedure very similar to the previous one. But instead of picking each
dimension to be reduced at random, dimension $i$
of the parameter
vector is chosen with probability 
$\frac{1}{i}$ (Zipf’s law). This
process now replicates the behavior of SGD training of MLPs very well:
initially, parameter vectors are wildly different, but they converge
towards a common local minimum while still retaining a memory of their
initial conditions over many training steps and many orders of
magnitude of the objective function.

\begin{figure}[thp]
 \includegraphics[width=\textwidth]{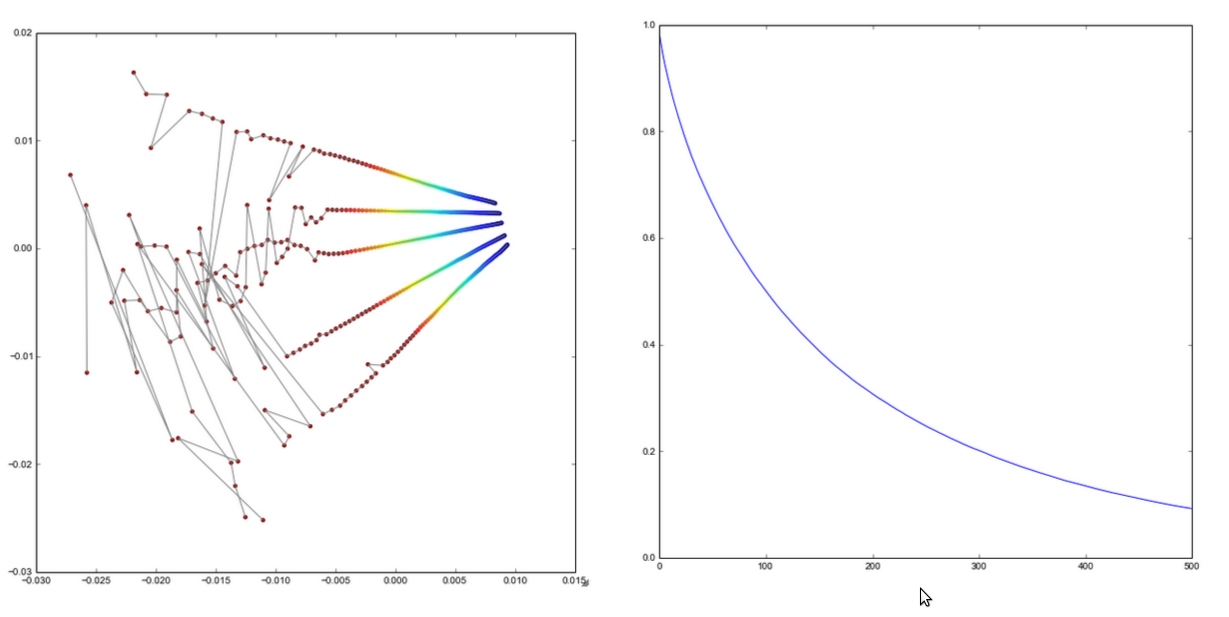} 
\caption{\label{zipf}
Simulated stochastic gradient descent of 5 random initial
1000-dimensional weight vector towards zero, after multidimensional
scaling. At each time step, a dimension is chosen at random according
to Zipf’s law and multiplied by 0.9
}
\end{figure}

How does this simple stochastic model actually relate to MLP training?
Notions of “local minima”, “valleys” and “basins of attraction” that
are usually used to justify training strategies for MLPs really imagine
the optimization problem to be one that makes complicated tradeoffs
between many different parameters in the optimization of a single
function. But we should actually think of the MLP training problem as
that of optimizing a large collection of largely independent and
unrelated functions. For example, the parameters that discriminate
between the digits 4/9 are largely unrelated to the parameters that
discriminate between the digits 0/8. Each such subtasks of the overall
classification problem simply gets optimized at its own rate, depending
on the frequency of relevant training examples. If training samples
required for learning a particular subtask of the problem are rare,
that subtask will converge slowly, while others converge fast.

\section{Discussion and Conclusions}

The experiments reported here suggest that the space in which SGD
optimization of neural network weights takes place has properties that
differ significantly from those usually assumed. In particular, the
experiments show no evidence of the existence of local minima in which
the optimization process gets trapped and they show no evidence of the
existence of structures at different scales that would require
decreasing the learning rate over time. \footnote{ We should note that
these results are only true for the kinds of large networks studied
here. Very small networks (a few hidden units) trained on problems like
approximate XOR networks do experience local minima.\par }

At the same time, these visualizations clearly show that networks retain
“memory” of their initial random parameter initializations over
hundreds of epochs and orders of magnitude of changes in outputs,
weights, and error rates.

A tentative explanation of these observations is that SGD optimization
of neural networks does not proceed as the optimization of a single
function in a single space, but as an optimization of many functions in
otherwise largely independent subspaces, where updates of different
subspaces happen at greatly different rates. In the stochastic model
used to illustrate this idea above, these independent subspaces were
aligned with the coordinate axes, but they might also simply represent
arbitrary linear or non-linear subspaces in weight space.

This view of weight space is somewhat related to ideas about
optimization along “saddle points” that has been explored in second
order learning methods for neural networks. However, if
optimization really happens at very different rates in different
subspaces, it points out another problem with second order methods: it
means that the estimate of the local second order structure of the
objective function in weight space will have very different variances
in different directions.

Although these results seem generally consistent with training of other
network architectures (provided they are sufficiently large) and
training on other data sets, they remain to be verified. In particular,
it is interesting to see how these results carry over to recurrent
neural networks and LSTMs.

A model of SGD neural network optimization that views this optimization
as optimization of largely independent subspaces at different rates has
potential implications for the construction of better training
algorithms; these will be explored in another paper.

\bibliographystyle{plain}
\bibliography{mlp-conv}

\end{document}